\newif\ifeusipstyle
\newif\ifdohybrid
\eusipstylefalse
\dohybridfalse

\ifeusipstyle
  \documentclass[a4paper]{article}
  \usepackage{spconf,amsmath,graphicx,cite}
\else
  \documentclass[conference]{IEEEtran} 

  \usepackage[dvips]{graphicx}
\fi

\usepackage{url}
\usepackage{amsfonts}
\usepackage{color}

\title{Using the Projected Belief Network at High Dimensions}

\ifeusipstyle
   \name{Paul M Baggenstoss}
   \address{Fraunhofer FKIE, Fraunhoferstr 20,
   \\ 53343 Wachtberg, Germany}
\else
   \author{\IEEEauthorblockN{Paul M. Baggenstoss}
   \IEEEauthorblockA{Fraunhofer FKIE,
   Fraunhoferstrasse 20\\
   53343 Wachtberg, Germany\\
   Email: p.m.baggenstoss@ieee.org}
   \and
   \IEEEauthorblockN{Frank Kurth}
   \IEEEauthorblockA{Fraunhofer FKIE,
   Fraunhoferstrasse 20\\
   53343 Wachtberg, Germany\\
   Email: frank.kurth@fkie.fraunhofer.de}
   }
\fi

\begin{document}
\newcommand{\defined}{\stackrel{\mbox{\tiny$\Delta$}}{=}}
\newtheorem{example}{Example}
\newtheorem{conclusion}{Conclusion}
\newtheorem{assumption}{Assumption}
\newtheorem{definition}{Definition}
\newtheorem{problem}{Problem}
\newcommand{\erf}{{\rm erf}}

\newcommand{\sst}{\scriptstyle }
\newcommand{\xparen}{\mbox{\small$(\bfx)$}}
\newcommand{\hojz}{H_{0j}\mbox{\small$(\bfz)$}}
\newcommand{\Hozj}{H_{0,j}\mbox{\small$(\bfz_j)$}}
\newcommand{\smallmath}[1]{{\scriptstyle #1}}
\newcommand{\Hoz}[1]{H_0\mbox{\small$(#1)$}}
\newcommand{\Hozp}[1]{H_0^\prime\mbox{\small$(#1)$}}
\newcommand{\Hozpp}[1]{H_0^{\prime\prime}\mbox{\small$(#1)$}}
\newcommand{\hoz}{\Hoz{\bfz}}
\newcommand{\hooz}{\Hozp{\bfz}}
\newcommand{\hoooz}{\Hozpp{\bfz}}
\newcommand{\smJ}{{\scriptscriptstyle \! J}}
\newcommand{\smK}{{\scriptscriptstyle \! K}}

\newcommand{\erfc}{{\rm erfc}}
\newcommand{\bitem}{\begin{itemize}}
\newcommand{\dsum}{{ \displaystyle \sum}}
\newcommand{\eitem}{\end{itemize}}
\newcommand{\benum}{\begin{enumerate}}
\newcommand{\eenum}{\end{enumerate}}
\newcommand{\bdm}{\begin{displaymath}}
\newcommand{\bfzro}{{\underline{\bf 0}}}
\newcommand{\bfone}{{\underline{\bf 1}}}
\newcommand{\edm}{\end{displaymath}}
\newcommand{\beq}{\begin{equation}}
\newcommand{\bea}{\begin{eqnarray}}
\newcommand{\eea}{\end{eqnarray}}
\newcommand{\cali}{ {\cal \bf I}}
\newcommand{\caln}{ {\cal \bf N}}
\newcommand{\barray}{\begin{displaymath} \begin{array}{rcl}}
\newcommand{\earray}{\end{array}\end{displaymath}}
\newcommand{\eeq}{\end{equation}}
\newcommand{\btheta}{\mbox{\boldmath $\theta$}}
\newcommand{\bTheta}{\mbox{\boldmath $\Theta$}}
\newcommand{\blam}{\mbox{\boldmath $\Lambda$}}
\newcommand{\beps}{\mbox{\boldmath $\epsilon$}}
\newcommand{\bdelta}{\mbox{\boldmath $\delta$}}
\newcommand{\bgamma}{\mbox{\boldmath $\gamma$}}
\newcommand{\balpha}{\mbox{\boldmath $\alpha$}}
\newcommand{\bbeta}{\mbox{\boldmath $\beta$}}
\newcommand{\balphascript}{\mbox{\boldmath ${\scriptstyle \alpha}$}}
\newcommand{\bbetascript}{\mbox{\boldmath ${\scriptstyle \beta}$}}
\newcommand{\bLambda}{\mbox{\boldmath $\Lambda$}}
\newcommand{\bDelta}{\mbox{\boldmath $\Delta$}}
\newcommand{\bomega}{\mbox{\boldmath $\omega$}}
\newcommand{\bOmega}{\mbox{\boldmath $\Omega$}}
\newcommand{\blambda}{\mbox{\boldmath $\lambda$}}
\newcommand{\bphi}{\mbox{\boldmath $\phi$}}
\newcommand{\bpi}{\mbox{\boldmath $\pi$}}
\newcommand{\bnu}{\mbox{\boldmath $\nu$}}
\newcommand{\brho}{\mbox{\boldmath $\rho$}}
\newcommand{\bmu}{\mbox{\boldmath $\mu$}}
\newcommand{\sigi}{\mbox{\boldmath $\Sigma$}_i}
\newcommand{\bfu}{{\bf u}}
\newcommand{\bfx}{{\bf x}}
\newcommand{\bfb}{{\bf b}}
\newcommand{\bfk}{{\bf k}}
\newcommand{\bfc}{{\bf c}}
\newcommand{\bfv}{{\bf v}}
\newcommand{\bfn}{{\bf n}}
\newcommand{\bfK}{{\bf K}}
\newcommand{\bfh}{{\bf h}}
\newcommand{\bff}{{\bf f}}
\newcommand{\bfg}{{\bf g}}
\newcommand{\bfe}{{\bf e}}
\newcommand{\bfr}{{\bf r}}
\newcommand{\bfw}{{\bf w}}
\newcommand{\calX}{{\cal X}}
\newcommand{\calZ}{{\cal Z}}
\newcommand{\bb}{{\bf b}}
\newcommand{\bfy}{{\bf y}}
\newcommand{\bfz}{{\bf z}}
\newcommand{\bfs}{{\bf s}}
\newcommand{\bfa}{{\bf a}}
\newcommand{\bfA}{{\bf A}}
\newcommand{\bfB}{{\bf B}}
\newcommand{\bfV}{{\bf V}}
\newcommand{\bfZ}{{\bf Z}}
\newcommand{\bfH}{{\bf H}}
\newcommand{\bfX}{{\bf X}}
\newcommand{\bfR}{{\bf R}}
\newcommand{\bfF}{{\bf F}}
\newcommand{\bfS}{{\bf S}}
\newcommand{\bfC}{{\bf C}}
\newcommand{\bfI}{{\bf I}}
\newcommand{\bfO}{{\bf O}}
\newcommand{\bfU}{{\bf U}}
\newcommand{\bfD}{{\bf D}}
\newcommand{\bfY}{{\bf Y}}
\newcommand{\bSig}{{\bf \Sigma}}
\newcommand{\test}{\stackrel{<}{>}}
\newcommand{\zmk}{{\bf Z}_{m,k}}
\newcommand{\zlk}{{\bf Z}_{l,k}}
\newcommand{\zm}{{\bf Z}_{m}}
\newcommand{\ssq}{\sigma^{2}}
\newcommand{\dint}{{\displaystyle \int}}
\newcommand{\ds}{\displaystyle }
\newtheorem{theorem}{Theorem}
\newcommand{\postscript}[2]{ \begin{center}
    \includegraphics*[width=3.5in,height=#1]{#2.eps}
    \end{center} }

\newtheorem{identity}{Identity}
\newtheorem{hypothesis}{Hypothesis}
\newcommand{\mathtiny}[1]{\mbox{\tiny$#1$}}

\maketitle

\begin{abstract}
The projected belief network (PBN) is a layered generative network (LGN)
with tractable likelihood function, and is based on a feed-forward
neural network (FFNN).  There are two versions of the PBN: stochastic and
deterministic (D-PBN), and each has theoretical advantages over other LGNs.   
However, implementation of the PBN requires 
an iterative algorithm that includes
the inversion of a symmetric matrix of size $M\times M$ in each layer, where $M$ is the layer
output dimension. This, and the fact that the network must be always
dimension-reducing in each layer, can limit the types of problems
where the PBN can be applied.  In this paper, we describe techniques
to avoid or mitigate these restrictions and use the PBN effectively at high dimension. 
We apply the discriminatively aligned PBN (PBN-DA)
to classifying and auto-encoding high-dimensional spectrograms of acoustic events.
We also present the discriminatively aligned D-PBN for the first time.
\end{abstract}

\section{Introduction}
\subsection{Motivation: Advantages of PBN}
The projected belief network (PBN) is a layered generative network (LGN)
with tractable likelihood function, and is based on a feed-forward
neural network (FFNN).  There are two versions of the PBN: stochastic and
deterministic, and each has  theoretical advantages over other layered generative networks.   

It has recently been shown that  
the information at the output of a dimension-reducing
transformation can be maximized
using probability density function projection (PDF projection) \cite{BagKayInfo2022}.
PDF projection estimates the distribution of the input data, simultaneously with a
dimension-reducing transformation that extracts the
latent variables \cite{BagPDFProj,Bag_info}. 
The method is more general than other methods of 
non-linear dimension reduction (NLDR) \cite{Rosenblatt,DecoHiOrder,DecoDragan,
OjaICA2001,HYVARINEN1999429}.  
Implementing PDF projection in a neural network architecture
is called projected belief network (PBN)
\cite{BagIcasspPBN,BagEusipcoPBN, BagPBNEUSIPCO2019, BagPBNEUSIPCO2020}.
As a result, the PBN is the most direct and general way to apply NLDR in a neural network architecture.

The LF of other widely used LGNs can only be obtained by intractable
integration (marginalizing) over the hidden variables.
They must rely on a surrogate cost function in order to approximate LF training.
Examples are contrastive divergence (CD) to train restricted 
Boltzmann machines \cite{WellingHinton04,HintonDeep06}, and Kullback
Leibler divergence to train variational auto-encoder (VAE) \cite{pmlr-v32-rezende14},
or an adversarial discriminative network to train generative adversarial networks
(GAN)  \cite{GoodfellowGAN2014}.  On the other hand, since the PBN generates data by manifold sampling, it
posesses a tractable likelihood function (LF) that allows direct gradient-based training.

The deterministic PBN (D-PBN) can be used as an auto-encoder
\cite{BagPBNEUSIPCO2019,BagIcasspPBN} and has theoretical advantage over conventional auto-encoders.
While other auto-encoders use an empirical
reconstruction network, the deterministic PBN reconstructs input data
by backing up (back-projecting) through the same feed-forward
neural network (FFNN) that was used to extract the features. In each layer, it selects
the conditional mean estimate of the layer input based on a maximum entropy prior,
a type of optimal estimator.

Another advantage of the PBN and D-PBN is that they are based on a FFNN.  Therefore, a
single FFNN can be simultaneously a generative and a discriminative
network \cite{BagPBNEUSIPCO2020}. This offers the most direct way to combine the 
advantages of both network types.  A number of variations of this concept have
been proposed.  The PBN or D-PBN cost functions can be used as a regularization for discriminative
neural networks \cite{BagPBNEUSIPCO2020}, or the opposite:
discriminative cost function can be used to ``align" a PBN
to decision boundaries to create better-performing generative models \cite{BagSPL2021}.
We will use this approach in this paper to test the PBN and D-PBN at high dimensions.

\subsection{Discriminative Alignment of PBN}
\label{sndbx}
It was shown in \cite{BagSPL2021} that a generative classifier
(a PBN) can compete with state of the art discriminative classifiers.
This seems to contradict the widely-held belief that the
generative task is much harder, and unnecessary for classifying \cite{Vapnik99}.
However, generative models are useful in their own right, and
can be applied in some tasks where discriminative classifiers
cannot \cite{Goodfellow2016}. A generative classifier than
can perform as well as a discriminative classifier,
is highly desireable.

{\bf Sand box analogy}.  Discriminative alignment can be understood in terms of the
conceptual image of a sand box enclosed by a wooden frame.  The frame can be
thought of as the discriminative task, separating the probability
mass (the sand) from the other classes at the decision boundaries.
The generative model can be thought of as heaping the sand
at the places where data is more likely.
The two tasks can be achieved simultaneously using
discriminative alignment of a PBN.
%

Despite the benefits it offers, the PBN
is bound by restrictions that also affect the discriminative
model that it may share : (a) dimension-reducing layers,  (b)
no max-pooling in convolutional layers, 
(c) no dropout regularization, and (d) no batch normalization.
On the positive side, L2 regularization can be used, and the generative
cost function can be seen as a form of regularization
in itself \cite{BagPBNEUSIPCO2020}, and one can avoid the restriction
on dimension-reducing layers using the method in Section \ref{ecsec}.

\subsection{Computational Challenge of PBN}
\label{cpch}
%
Despite these theoretical advantages, implementation of the PBN requires in each layer
the inversion of a $M\times M$ symmetric matrix where
$M$ is the layer output dimension.  In convolutional
layers, $M$ can be extremely large because it is the
product of the number of kernels with the size of the feature maps.
Depending on the type of PBN layer (i.e depending
in the type of maximum entropy prior distribution),
this matrix may need to be inverted for each sample in a mini-batch.

\subsection{Paper Organization and Contributions}
We begin by a mathematical review of PBN and D-PBN
in Section \ref{revsec}, explaining the computational challenge
of PBN at high dimensions, then suggest new approaches to applying PBN at high dimensions 
in Section \ref{mitisec}. Experimental results
are provided in Section \ref{expr} including a novel variation of PBN, called D-PBN-DA 
in Section \ref{dpbnda}. 

\section{The PBN: Review}
\label{revsec}

\subsection{PDF Projection}
A PBN is just a layer-wise application of PDF projection.
In probability density function (PDF) projection, one defines an $N\times 1$ input vector
$\bfx$, and a transformation $\bfz=T(\bfx)$ producing
the $M\times 1$ output vector $\bfz$, where $M<N$.
Suppose we have data $\bfx$ drawn from some distribution
$p(\bfx)$ with support on $\mathbb{X} \subseteq \mathbb{R}^N$, that we'd like to estimate.  
Although $p(\bfx)$ is unknown, we know $g(\bfz)$, which is an estimate of the feature distribution
of $\bfz$, where $\bfx$ is drawn from $p(\bfx)$ and $\bfz=T(\bfx)$, and
has support on $\mathbb{Z}\subseteq \mathbb{R}^M$.
To estimate $p(\bfx)$ based on knowing $g(\bfz)$, we first define a  reference (or prior) distribution $p_0(\bfx)$
for which we know $p_0(\bfz)$, the mapping of $p_0(\bfx)$ through $T(\bfx)$.
Then, the PDF projection theorem \cite{BagPDFProj} states that
the function 
\beq
G(\bfx;T,p_0,g)=\frac{p_0(\bfx)}{p_0(\bfz)} \; g(\bfz),
\label{ppt0}
\eeq
where $\bfz=T(\bfx)$, is a PDF on $\mathbb{X}$ (it integrates to 1) and is among the class of PDFs which
map to distribution $g(\bfz)$ through $T(\bfx)$.
Simply stated, $G(\bfx;T,p_0,g)$ is an estimate of $p(\bfx)$ based on
prior distribution $p_0(\bfx)$ and feature PDF estimate $g(\bfz)$.

The biggest challenge on PDF projection is the derivation of $p_0(\bfz)$.
When $p_0(\bfx)$ is a canonical reference distribution,
the feature distribution $p_0(\bfz)$ is known in closed form
for many types of transformations \cite{BagNutKay2000}. 
For an even broader class of reference distributions and transformations,
$p_0(\bfz)$ is not known, but the moment generating function (MGF)  
is known exactly. In these cases, we can invert the MGF using the
saddle point approximation (see eq. (16) in \cite{BagNutKay2000}). 
We note that the term ``approximation" is misleading
because the error can be neglected for all practical purposes
\cite{BagSPL2021}.  

\subsection{The PBN}
For complex transformations, such as a neural network,
$p_0(\bfz)$ may be difficult or impossible to derive,
so we apply PDF projection recursively (layer-wise).
This results in a projected belief network (PBN) \cite{BagPBN},
where the likelihood function (LF) $G(\bfx;T,p_0,g)$ becomes a
product of terms, each term being the application of (\ref{ppt0})
to a layer (see eq. (1) in \cite{BagPBNEUSIPCO2020}).

\subsection{Finding the Saddle Point}
Each term in the LF of a PBN requires computing the saddle point.
Let $\bfx\subseteq\mathbb{R}^N$ be the input of a layer, where
$N$ is the dimension.  Let $\bfz\subseteq\mathbb{R}^M$ be the output of the linear
transformation given by $\bfz={\bf W}^\prime \bfx$.
This is the dimension-reducing linear transformation used in each layer of
a neural network before application of the activation function.
Note that this equation holds both for dense (fully connected) 
and convolutional layers. Although convolutional layers are implemented
by efficient convolution, there exists at least theoretically
an $N\times M$ matrix ${\bf W}$.

The saddle point $\hat{\bfh}$ is the $M\times 1$ vector $\bfh$ that solves
\beq
{\bf W}^\prime \lambda\left(\balpha_0 + {\bf W} \bfh\right) = \bfz,
\label{tm1}
\eeq
where 
$N\times 1$ constant vector $\balpha_0$ and element-wise activation function $\lambda(\;)$ 
depend on the chosen MaxEnt prior \cite{BagIcasspPBN}.

Equation (\ref{tm1})  is solved by an iterative algorithm (Newton-Raphson)
that involves inverting the matrix ${\bf C}(\hat{\bfh})={\bf W}^\prime \Lambda {\bf W}$,
where $\Lambda$ is the $N\times N$ diagonal matrix with diagonal
$\lambda^\prime(\alpha_1), \lambda^\prime(\alpha_2), \ldots \lambda^\prime(\alpha_N)$, where
$\lambda^\prime(\alpha)$ is the value of the first derivative of
$\lambda\left(\alpha\right)$ with repect to $\alpha$, 
and  $\balpha=\balpha_0+{\bf W} \hat{\bfh}$.

\subsection{Using the Saddle Point}
The deterministic PBN (D-PBN) reconstructs $\bfx$ from $\bfz$  and 
the saddle point as follows  \cite{BagIcasspPBN}: 
\beq
\hat{\bfx}= \lambda\left(\balpha_0 + {\bf W} \hat{\bfh}\right) 
\label{cm1}
\eeq
A multi-layer  D-PBN is created by propagating the estimate backward in the network \cite{BagIcasspPBN,BagPBNEUSIPCO2019}.
The stochastic PBN uses the saddle point to compute terms in the LF.
Specifically, $p_0(\bfz)$ is written using equation (16) in \cite{BagNutKay2000},
where the saddle point is written $\hat{\blambda}$ and $p_0(\bfz)$ is written $p_z(\bfu)$.

\section{Dimension Mitigation}
\label{mitisec}
The computational challenges of using PBN at high dimensions,
outlined in Section \ref{cpch}, can be mitigated using the following approaches.

\subsection{Seamless Streaming and Reconstruction}
Large data samples that have a time dimension, such as audio
recordings can be broken into smaller segments and processed 
by a sliding window.  There is an implicit assumption of independence 
between segments here that can be approximately true if the segments are large 
enough.  As long as the sliding window is wide enough to capture
enough temporal information, the generative model can provide 
meaningful and significant dimension reduction with high information 
content.  The collection of output features, i.e. the ``stream"
of output features can be used to reconstruct the
complete event, and further feature dimension reduction can be 
done on the assembled stream so that temporal information
spanning larger time durations can be extracted downstream.

For seamless reconstruction, the sliding windows must overlap by exactly 
the right amount, so that the assembled output stream is no different
than what would have been created by the same processing on the entire event.
Any temporal processing, i.e. temporal convolution, must be done without edge 
effects.  For this reason, convolutions in convolutional network layers must be done
without zero-padding.

\subsection{Gaussian Layers}
A very significant speed-up is afforded by Gaussian layers.
In a Gaussian layer, the Gaussian MaxEnt prior is
assumed, and $\lambda(x)=x$ is the linear activation,
meaning that ${\bf C}(\hat{\bfh})={\bf C}={\bf W}^\prime{\bf W}$ is independent of $\hat{\bfh}$
and  $\balpha_0={\bf 0}$, so  (\ref{tm1}) can be solved in closed form,
i.e.  $\hat{\bfh}=({\bf W}^\prime{\bf W})^{-1} \bfz,$
and needs to be done only once per mini-batch.
Note that for convolutional layers, all operations
are done by convolution, and matrix ${\bf W}$ never needs
to be actually created.

Using Gaussian layers, however, assumes that the layer input 
has support on $\mathbb{R}^N$, which means that
 activation functions should be avoided in the layer that precedes 
a Gaussian layer to attain a good PDF estimate using PDF projection.
Using linear activations in the first few layers, and
then non-linear activations in middle and low-dimensional
layers may be an effective compromise.

\subsection{Gaussian Layer Groups (GLG)}
For very large hidden variable dimensions, using
a Gaussian layer may be prohibitive - even allocating
memory for ${\bf C}$ may be difficult.  In these cases, 
it is possible to use several Gaussian layers
in a row, ending in a manageable output dimension,
then grouping the linear transformations
so that they are representable by a single linear transformation with low
output dimension $M$, and small ${\bf C}$.
After the GLG, non-linear activation functions can be used.
As long as $M$  at the group output is large enough, significant
functional approximation power is afforded by the network.


\subsection{Dimension-Preserving Layers}
If a layer input and output have the same dimension,
it is a dimension-preserving (1:1) transformation layer
and computing the saddle point is not required. It is only necessary
to solve for the determinant of the Jacobian matrix of the
transformation as a whole. 

\subsection{Expand-Contract Groups (ECG)}
\label{ecsec}
To afford even greater functional approximation power, it is possible to
expand the hidden variable dimension using a series of 
high-dimensional network layers, then reduce the dimension back down
to what it was at the start. This group of layers, seen as a unit,
can be analyzed as a dimension-preserving (1:1) transformation,
as if it was a single dimension-preserving layer.

\subsection{Direct Saddle Point Estimation}
One can find $\hat{\bfh}$, the solution to (\ref{tm1}),
by direct inversion using a neural network by training
on sample pairs $\bfz$, $\hat{\bfh}$. Error in the result is removed by a few iterations of
the Newton-Raphson algorithm.

\section{Experimental Results}
\label{expr}


\subsection{Experimental Goal: Discriminative Alignment of PBN}
The goal of our first experiment was to show that
discriminative alignment, which was demonstrated previously
on input data data dimensions of 196 and 900 \cite{BagSPL2021}, 
can be used on a different data set with much higher dimension.
Discriminative alignment is explained in Section \ref{sndbx}
and further details are given in Section \ref{trnsec}.

\subsection{Data and Feature Extraction}
The Office Sounds database \cite{OfficeSounds,BagEusipcoMRHMM2018} contains
twenty-four signal classes containing 102 samples
of each class, a total of 2448 example sounds.
Most of the sounds are created by
dropping common objects or operating office tools such as scissors
or staplers.  All time-series are 16128 samples long
(1/2 second in duration at 32000 Hz). 
We selected six of the twenty-four classes that had the highest
inter-class errors.  These classes were ``skit",  ``sciss", ``pret", ``pens", ``paper", and ``jing"
and are seen on the top row of Figure \ref{dpbnrecon}.
There were 102 samples of each class, so 612 samples total.
The emphasis is to achieve very high correct
classification performance on acoustic data with few training samples.
The 612 events were divided into training and testing sets of 306 each.
We conducted two-fold hold-out in which we trained on half of the data and tested using the other half,
then repeated the experiment with the data switched.
To extract features, we segmented each time-series sample into size-384 windows with 2/3 overlap,
producing 126 segments per time series.  We then took the FFT, magnitude-squared, then the log.
The features were therefore $126\times 193$, or dimension 24,318.

\subsection{Network, Algorithm, and Hardware}
The network consisted of seven layers:
\benum
  \item Convolutional with 6 ($9\times 10$) kernels, with   ($3\times 3$) downsampling, linear activation,
  \item Convolutional with 36 ($7\times 8$) kernels, with   ($3\times 3$) downsampling, linear activation, 
  \item Convolutional with 96 ($8\times 7$) kernels, with   ($2\times 3$) downsampling, linear activation,
  \item Dense with 512 neurons, linear activation. 
  \item Dense with 256 neurons, TG activation. 
  \item Dense with 128 neurons, TG activation. 
  \item Classifier layer, dense with 6 neurons. 
\eenum

The ``TG" activation is the truncated Gaussian activation \cite{Bag2021ITG}, similar in behavior to softplus, not unlike
leaky Relu, but continuous. All convolutions used the ``valid" border mode,
so no zero-padding was used and the output feature maps are smaller than the input
map, even without down-sampling.
The output feature maps of the five layers have shapes (6, 40, 62),  (36, 12, 19), (96, 3, 5), 512, 256, 128, and 6,
with total dimensions 14880, 8208, 1440, 512, 256, 128, and 6 respectively.

\subsection{Dimension Mitigation and Training}
Seamless streaming was not neccessary for this data set because the entire
event of dimension 24,318 could be processed.
However, note that the first five layers form a GLG with output dimension
256, so the saddle point can be solved by inverting a matrix ${\bf C}$ of only $256\times 256$,
just once per mini-batch. 
Interestingly, the dominant computational load in the network involves the
6-th layer, which requires inverting a $128\times 128$ matrix ${\bf C}(\hat{\bfh})$ 
on each sample of the mini-batch.
The algorithm maximized the log-likelihood of the PBN.
We used direct saddle point estimation using a 
symmetric linear transformation, followed by some Newton-Raphson iterations.
Surprisingly, a linear transformation $\hat{\bfh} = {\bf A}^\prime {\bf A} \bfz$,
results in a small residual error.  
We used a mini-batch size of 306, which was all of the training data.
The large batch size made best use of the GPU.
On a Quadro P6000 GPU with 32 bit (single precision) we were able
compute an epoch in 8.5 seconds, or 11 seconds with 64 bit precision.
All experiments were done using the PBN Toolkit, which is a graphical
tool using the Theano framework.  Each model required about 500 epochs. 


\subsection{Benchmark Network}
For a performance benchmark, we applied conventional deep convolutional neural network 
(CNN) with state of the art training methods.  We used the same network structure as the PBN, but the CNN was not bound by the restrictions of the PBN.
Therefore, we used TG activation function in each layer (linear activation was not used), and
we used max-pooling in the convolutional layers (not down-sampling), dropout, and L2 regularization.  

\subsection{PBN and DNN Training}
\label{trnsec}
A separate PBN was trained for each class by maximizing a combined generative-discriminative target function
consisting of (a) the PBN log-likelihood function, and (b) the negative cross-entropy classifier cost for discriminating 
the corrsponding class from all other classes.  The first component was given a weight of one for all samples from the target class, and zero
for all samples from other classes.  The second component was weighted by one for all classes.
This strategy caused the network to train as a PBN for the target class, and as a classifier
against all other classes.  
Data was classified by  evaluating the PBN log-likelihood function for each class assumption, and choosing the largest.
A class-dependent prior density was used for the output distribution, as explained in \cite{BagSPL2021}.
A benchmark DNN was also trained using the same data partitioning.

\subsection{Classifier Results}
Figure \ref{pbnc_os3r_comb} shows the results of the classifier combination experiment
in which  the PBN log-likelihood and DNN output were combined with a variable combing factor.
On the far left is seen the DNN performance, and on the far right the 
discriminatively aligned PBN (PBN-DA) performance. 
The two classifiers alone had exactly the same error probability.  In the center of the graph, a deep drop in classifier error is seen,
reaching about half the error of each classifier individually.
Classifier combining works best when (a) the two classifiers have comparable
performance and (b) the classifiers are based on independent views of the data,
what usually occurs when combining generative and discriminative classifiers.
This result reflects what was seen on other data sets for PBN-DA (See Figures 4 and 5 in \cite{BagSPL2021}).
\begin{figure*}[ht!]
  \begin{center}
    \includegraphics[width=6.5in]{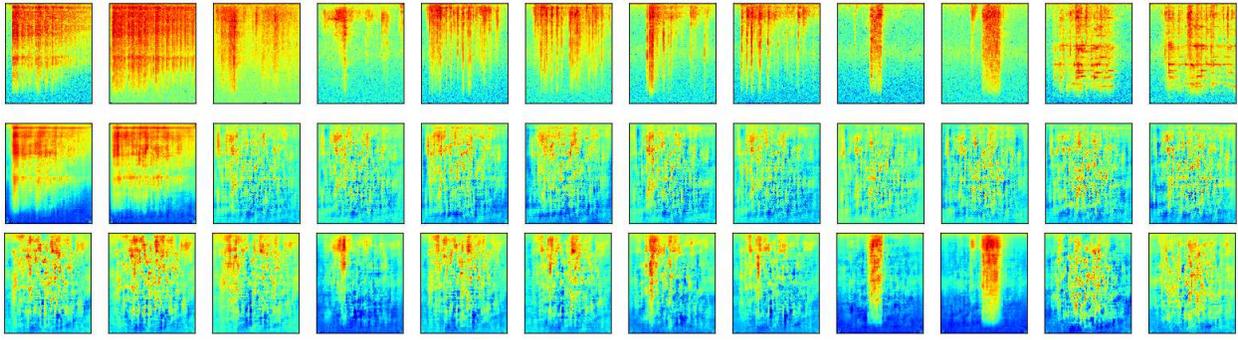}
  \caption{Original (top) and reconstructed (center and bottom) spectrograms using D-PBN. 
 There are two samples of each class in the order ``skit",  ``sciss", ``pret", ``pens", ``paper", and ``jing".
 A description of the sounds is found in \cite{OfficeSounds}. As can be clearly seen, the two rows of reconstructed samples
were trained on the first and fifth data class.}
  \label{dpbnrecon}
  \end{center}
\end{figure*}
\begin{figure}[h!]
  \begin{center}
    \includegraphics[width=3.0in, height=1.3in]{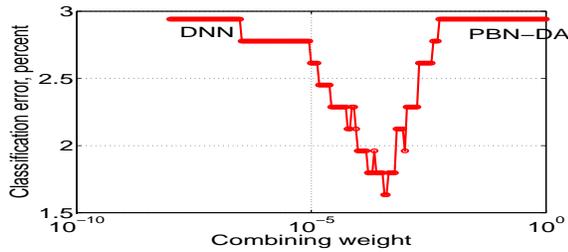}
  \caption{Classifier combination experiment. At the left
side, performance with just DNN, and on the right side just PBN-DA.}
  \label{pbnc_os3r_comb}
  \end{center}
\end{figure}

\subsection{D-PBN-DA}
\label{dpbnda}
To demonstrate the application of deterministic PBN (D-PBN) auto-encoder at high dimension,
we trained D-PBN on individual classes by reconstructing the input data from the sixth-layer output (256 dimension)
as explained in \cite{BagIcasspPBN}. There is a 100-fold reduction in dimension.   In contrast to previous uses of D-PBN, we applied a 
discriminative component to the cost function, exactly as done for the PBN-DA.  
 Figure \ref{dpbnrecon} shows examples of reconstructed samples.  On the top are a set of 12 original events, two from each class.
Below this are seen the reconstructions from the models
trained on the first and fifth class. The class correspondence can be easily
seen because only data from the target class is well reconstructed.  One can classify based on this reconstruction selectivity 
by choosing the class with lowest reconstruction error.  The classifier confusion matrix is 
(in errors out of 612 samples):
{
\begin{verbatim}
 [102  0   0   0   0   0]
 [ 0  90   1   1  10   0]
 [ 0   2  94   4   2   0]
 [ 0   1   1 100   0   0]
 [ 0   1   2   0  99   0]
 [ 0   0   0   0   0  102]
\end{verbatim}
}
having 25 errors in 612 samples (4.08\% error), almost as good as PBN-DA and DNN.
Due to the excellent class selectivity in reconstruction error,
the D-PBN-DA has promising potential applications in open-set classification.

\section{Conclusions}
Using some of the dimension mitigation methods proposed in Section \ref{mitisec},
we were able to effectively apply PBN at an input data dimension of 24,318, much higher
than the previously demonstrated dimensions of 900 and 196 (See Figures 4 and 5 in \cite{BagSPL2021}).
Furthermore, Figure \ref{pbnc_os3r_comb} confirms that the advantages of 
a discriminatively aligned PBN seen at lower dimensions, (a) PBN-DA working about as well as a state of the art DNN,
(b) the combined performance being significantly better that either classifier alone,
also apply at higher dimension.  In addition, we have demonstrated a novel new auto-encoder classifier (D-PBN-DA)
that performs as well as a state of the art CNN on the same high-dimensional data set.
The data, software, and instruction for recreating these results are archived at \cite{PBNTk}.

\bibliographystyle{ieeetr}
\bibliography{ppt}
\end{document}